\documentclass[sigconf]{acmart}
\AtBeginDocument{%
  \providecommand\BibTeX{{%
    \normalfont B\kern-0.5em{\scshape i\kern-0.25em b}\kern-0.8em\TeX}}}

\copyrightyear{2022}
\acmYear{2022}
\setcopyright{acmcopyright}\acmConference[MM '22]{Proceedings of the 30th ACM International Conference on Multimedia}{October 10--14, 2022}{Lisboa, Portugal} \acmBooktitle{Proceedings of the 30th ACM International Conference on Multimedia (MM '22), October 10--14, 2022, Lisboa, Portugal}
\acmPrice{15.00}
\acmDOI{10.1145/3503161.3548398}
\acmISBN{978-1-4503-9203-7/22/10}

\usepackage{algorithm}
\usepackage{algorithmic}
\usepackage{amsmath}
\usepackage{enumitem}
\usepackage{url}

\begin{document}

\title{Leveraging GAN Priors for Few-Shot Part Segmentation}



 \author{Mengya Han$^{*}$}
 \affiliation{
   \institution{Wuhan University}
   \city{Wuhan}
   \country{China}}
 \email{myhan1996@whu.edu.cn}

 \author{Heliang Zheng}
 \affiliation{%
   \institution{JD Explore Academy}
   \city{Beijing}
   \country{China}}
 \email{zhengheliang@jd.com}

 \author{Chaoyue Wang}
 \affiliation{%
\institution{JD Explore Academy}
   \city{Beijing}
   \country{China}}
\email{wangchaoyue9@jd.com}

 \author{Yong Luo$^{\dag}$}
 \affiliation{
   \institution{Wuhan University \&  \\Hubei Luojia Laboratory}
   \city{Wuhan}
   \country{China}}
 \email{luoyong@whu.edu.cn}
 
 \author{Han Hu}
 \affiliation{
   \institution{Beijing Institute of Technology}
   \city{Beijing}
   \country{China}}
 \email{hhu@bit.edu.cn} 
 
 \author{Bo Du$^{\dag}$}
 \affiliation{
   \institution{Wuhan University \& \\ Hubei Luojia Laboratory}
   \city{Wuhan}
   \country{China}}
 \email{dubo@whu.edu.cn}
 
 \thanks{* This work was performed when Mengya Han was visiting JD Explore Academy as a research intern.}
\thanks{\dag Corresponding author: Yong Luo, Bo Du.}





\renewcommand{\shortauthors}{Mengya Han et al.}

\begin{abstract}

Few-shot part segmentation aims to separate different parts of an object given only a few annotated samples. Due to the challenge of limited data, existing works mainly focus on learning classifiers over \textbf{pre-trained features}, failing to learn \textbf{task-specific features} for part segmentation. In this paper, we propose to learn task-specific features in a ``pre-training''-``fine-tuning'' paradigm. We conduct prompt designing to reduce the gap between the pre-train task (\textit{i.e.}, image generation) and the downstream task (\textit{i.e.}, part segmentation), so that the GAN priors for generation can be leveraged for segmentation. This is achieved by projecting part segmentation maps into the RGB space and conducting interpolation between RGB segmentation maps and original images. Specifically, we design a fine-tuning strategy to progressively tune an image generator into a segmentation generator, where the supervision of the generator varying from images to segmentation maps by interpolation. Moreover, we propose a two-stream architecture, \textit{i.e.}, a segmentation stream to generate task-specific features, and an image stream to provide spatial constraints. The image stream can be regarded as a self-supervised auto-encoder, and this enables our model to benefit from large-scale support images. Overall, this work is an attempt to explore the internal relevance between generation tasks and perception tasks by prompt designing. Extensive experiments show that our model can achieve state-of-the-art performance on several part segmentation datasets.

\end{abstract}



\begin{teaserfigure}
  \centering
  \includegraphics[width=\textwidth]{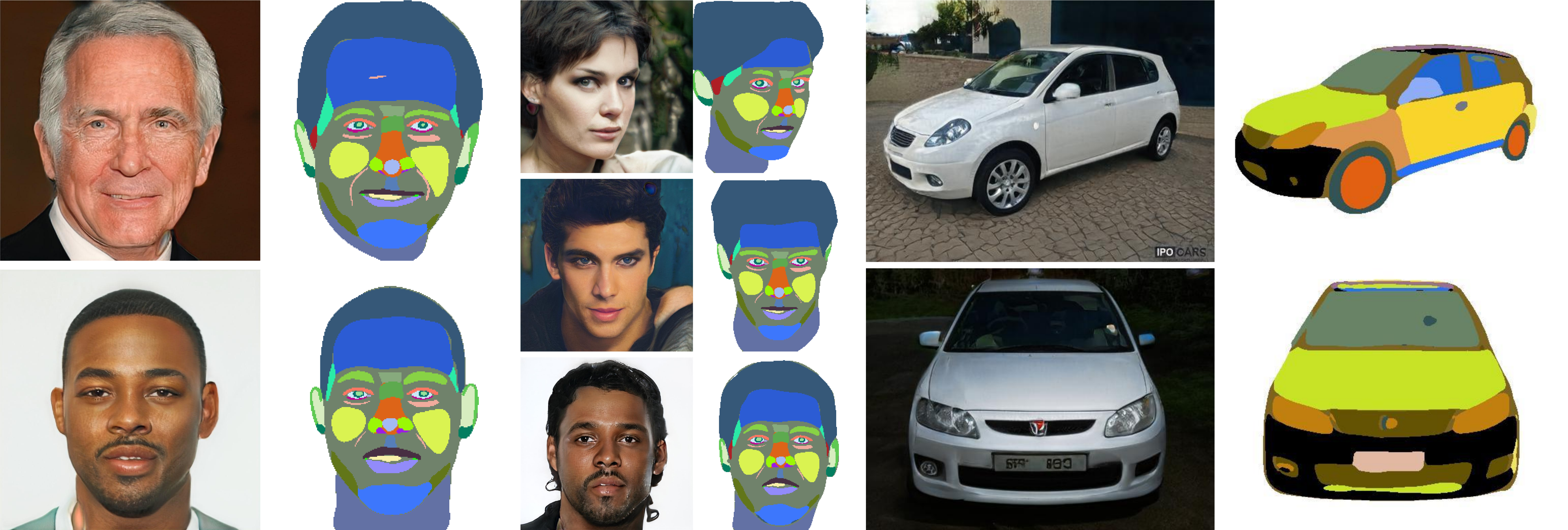}
  \caption{One-shot segmentation results. In each task, our segmentation network is given only one example of part labels.}
  \label{fig:teaser}
\end{teaserfigure}

\maketitle

\section{Introduction}



Few-shot part segmentation is an important task in computer vision that aims to separate different parts of an object given only a few annotated samples~\cite{tritrong2021repurposing,zhang2021datasetgan,baranchuk2021label,saha2021ganorcon}. In particular, the real world is composed of objects, which consist of different parts. Therefore, part-based image representation is a fundamental representation method that fits the inherent properties of objects~\cite{fischler1973representation, felzenszwalb2009object, hinton2021represent, he2021partimagenet, yang2020heterogeneous, DBLP:conf/cvpr/ChenMLFUY14}. However, due to the tremendous cost of labeling parts, there are still no large-scale datasets containing part labels, and hence it is extremely challenging to obtain part information. Existing work mainly focus on unsupervised/weakly-supervised part learning, while the lack of part-level supervision makes the training process unstable, and only coarse-grained semantic parts (\textit{e.g.}, 3-5 semantic parts for each object) can be obtained~\cite{two-attention, choudhury2021unsupervised, Part-align}. Recently, few-shot part segmentation has attracted an increasing attention~\cite{tritrong2021repurposing,zhang2021datasetgan,baranchuk2021label,saha2021ganorcon,ling2021editgan}. By involving only a few human efforts, such a setting significantly boosts part learning performance (\textit{e.g.}, the learned model is able to localize semantic 20-30 parts for each object), making it of great potential to be an effective way to obtain part-based representation.

Current few-shot part segmentation works mainly focus on utilizing and studying the impact of pre-trained features~\cite{tritrong2021repurposing,zhang2021datasetgan,baranchuk2021label,qiu2021synface,8627945}. Rep-GAN~\cite{tritrong2021repurposing} and DatasetGAN~\cite{zhang2021datasetgan} are some early works on few-shot part segmentation, and they extract pixel-wise representations from a pre-trained GAN and use them as feature vectors for a segmentation network. The experiments show that such pre-trained GAN features are ``readily discriminative'' and can produce surprisingly good results. GanOrCon~\cite{saha2021ganorcon} revisits these two works and argues that self-supervised models trained by contrastive learning can provide better features. DDPMS~\cite{baranchuk2021label} takes one step further and shows that pre-trained denoising diffusion probabilistic models can provide the most effective features for few-shot part segmentation and outperform existing baselines by a clear margin.

Although significant progresses have been achieved by utilizing \textbf{pre-trained features}, characteristics of the target task cannot be exploited. This limits the room for performance improvements of these approaches. As feature learning always matters in computer vision tasks, it is necessary to study the problem of learning \textbf{task-specific features} for part segmentation. However, it is non-trivial to use only a few samples to train or fine-tune a feature extractor, which is pre-trained over a large-scale dataset~\cite{karras2019style, yu2015lsun}. For example, training traditional semantic segmentation model usually requires thousands of annotations~\cite{Cordts2016Cityscapes, zhou2017scene}, and StyleGAN-ADA~\cite{karras2020training} finds that reducing the amount of data (FFHQ~\cite{karras2019style}) to 7k images would cause a severe overfitting in GAN training. Consequently, it is extremely challenging to learn task-specific features for part segmentation using only 1-5 samples.

To address this challenge, we propose a Progressive Fine-Tuned GAN (PFTGAN), which can learn task-specific features for part segmentation in a ``pre-training''-``fine-tuning'' paradigm. In particular, we train GANs to synthesis high-fidelity images in an unsupervised manner. In the synthesis process, the semantic priors about the object and part can be learned, and we propose to preserve and leverage such priors for part segmentation. Thus we project part segmentation maps into the RGB space. As a kind of prompt learning~\cite{zhou2021learning}, such a projection converts the original pixel-wise classification task into a RGB value regression (generation) task. Then a model for the pre-trained task (\textit{i.e.}, image generation) can be gradually tuned to complete the downstream task (\textit{i.e.}, few-shot part segmentation).

Specifically, our PFTGAN consists of an image encoder and a two-stream StyleGAN-based decoder. The encoder is a pre-trained GAN inversion model, which can transform a given image to the corresponding StyleGAN latent code. The decoder consists of a segmentation stream and an image stream, where the segmentation stream can generate task-specific features for part segmentation and the image stream can provide spatial constraints. Note that during training, the segmentation stream is optimized by few samples (\textit{i.e.}, RGB part segmentation maps), while the image stream can be regarded as a self-supervised auto-encoder and plenty of training images are available. The projection of part segmentation maps (to the RGB space) reduces the ``task gap'' between generation and segmentation. To further reduce the ``domain gap'' of RGB segmentation maps and original images, we design a fine-tuning strategy to progressively tune the segmentation stream from an image generator into a segmentation generator. This is achieved by conducting interpolation over RGB segmentation maps and images, where supervision of the segmentation stream gradually changes from original images to RGB segmentation maps. 

This work is an attempt to explore the internal relevance between generation tasks and perceptual tasks by prompts designing. Our main contributions can be summarized as follows:

\begin{itemize}[leftmargin=*]
\item We address the challenges of learning task-specific features for few-shot part segmentation by proposing a progressive fine-tuning strategy, which can effectively reduce the task gap and domain gap of image generation and part segmentation.
\item We propose a two-stream architecture where the self-supervised auto-encoder enables our model to benefit from large-scale support images and facilitate to preserve spatial information.
\item We conduct comprehensive experiments and draw \textbf{two conclusive insights} to the few-shot part segmentation community: 1) GAN-based models can significantly outperform self-supervised models and diffusion-based models when the training samples are extremely limited (\textit{e.g.}, one-shot), and 2) our proposed fine-tuning strategy can consistently improve part segmentation performance with a clear margin.
\end{itemize}



\begin{figure*}
  \includegraphics[width=\textwidth]{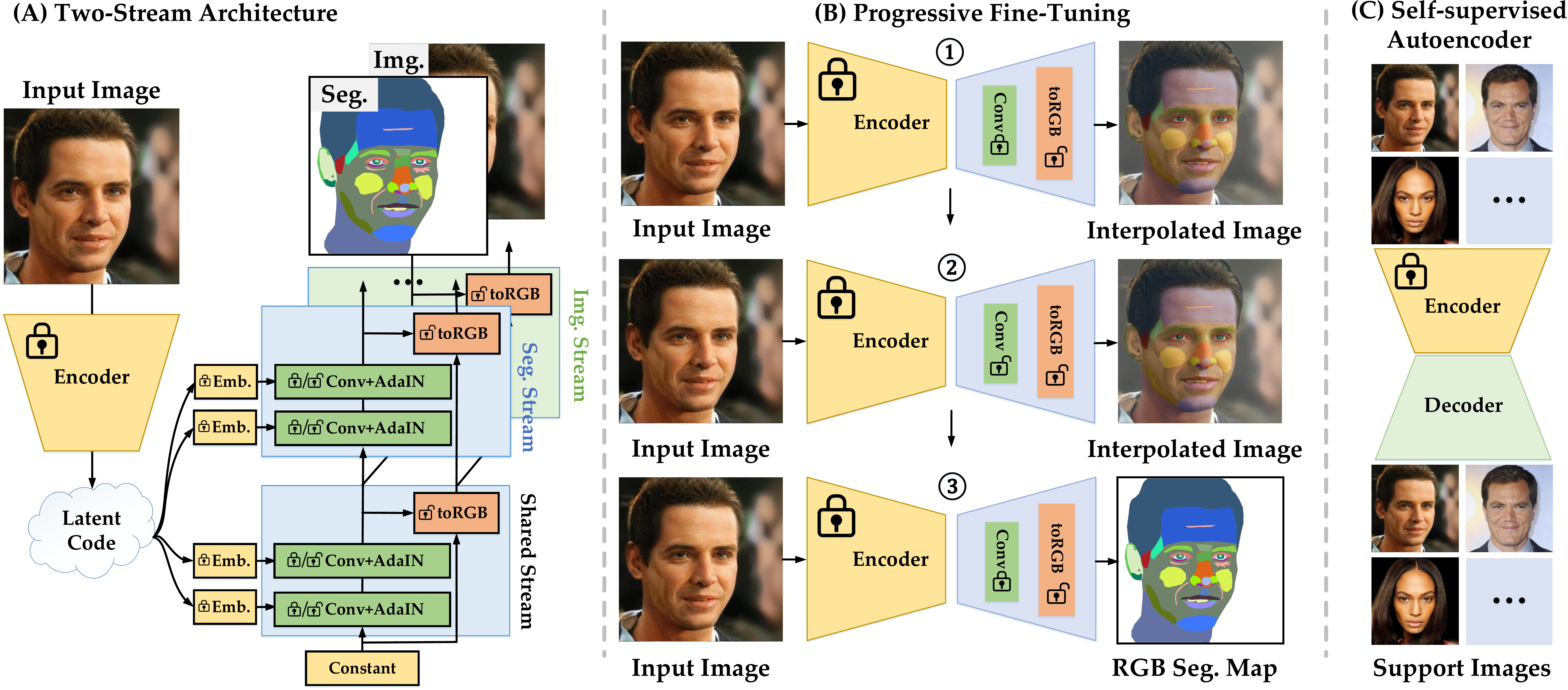}
  \caption{An overview of the proposed PFTGAN. (A) shows the architecture of PFTGAN, which consists of an encoder and a two-stream decoder. The two streams are supervised by RGB segmentation maps and original images, respectively, where the layers corresponding to coarse spatial resolutions (\textit{e.g.}, $4 \times 4$ to $32 \times 32$) are shared; (B) is an illustration of our proposed progressive fine-tuning for the segmentation stream. We first fix the parameters of convolutional layers and fine-tune ``toRGB'' layers using interpolated images. After that, we set convolutional layers to be trainable and fine-tune the decoder using interpolated images. Finally, we further fine-tune ``toRGB'' layers using RGB segmentation maps; (C) shows the optimization of the image stream, which can be regarded as a self-supervised auto-encoder and is trained on large-scale support (unlabeled) images.}
  \label{fig:framework}
\end{figure*}

\section{Related Work}
In this section, we briefly describe the existing lines of research relevant to our work. 

\subsection{Deep Generative Models}

Deep generative models have shown promising results in modeling image distribution, thus enabling the synthesis of realistic images. There are several classes of visual generative models which include autoencoders based on encoder-decoder architectures such as VAE~\cite{kingma2013auto} and its variants~\cite{van2017neural,razavi2019generating,lu2021discriminator}, autoregressive models~\cite{van2016conditional,chen2018pixelsnail}, generative adversarial networks (GANs)~\cite{goodfellow2014generative}, and diffusion models~\cite{song2020improved,ho2020denoising, dhariwal2021diffusion}. In summary, these generative models learn the distribution of real-world data (\textit{i.e.}, images) in an implicit or explicit way. In recent years, GANs achieve great success in image generation~\cite{salimans2016improved,arjovsky2017towards,arjovsky2017wasserstein,gulrajani2017improved,huang2021unifying,8627945,hu2020aesthetic,li2021discovering,li2022comprehensive}. 

In general, for most visual generative models, the ultimate goal is to generate high-quality visual contents that follow the training data distribution. Accompanying the rapid development of generative models, an interesting question was raised, \textit{i.e.}, `Does the well-trained generative model understand the real-world data?'. To answer this question, several works focus on exploring the perception, understanding capabilities of the pre-trained generative model. For example, it has been shown that rich and controllable semantics can be obtained in the latent space of pretrained generators~\cite{DBLP:journals/tog/AbdalZMW21, DBLP:conf/cvpr/ShenGTZ20, DBLP:conf/eccv/ZhuSZZ20, DBLP:journals/corr/abs-2107-13812}, and learning an encoder for a pretrained generator can obtain remarkable features that are generalizable across a wide range of computer vision applications~\cite{DBLP:conf/cvpr/XuSZYZ21}. DatasetGAN~\cite{zhang2021datasetgan} aims to learn the mapping function from the generated features to spatial semantic information. These methods have demonstrated that the generative model, which is capable of generating high fidelity images, should also perceive and understand the data distribution/semantic properties to some extent. In this paper, we further explore how the generative priors could help the perceiving model for improving part segmentation capability given very limited supervision.

\subsection{Few-shot Part Segmentation}
Past research has attempted to solve segmentation with few annotations~\cite{dong2021abpnet,liu2020dynamic,zhuge2021deep}. Few-shot part segmentation aims to recognize each part of an object in the given images according to only a few numbers of images with predefined part annotations~\cite{tritrong2021repurposing,zhang2021datasetgan,baranchuk2021label,saha2021ganorcon}. Recently, tremendous progress has been made in this field, benefiting from the advantages of generative models~\cite{tritrong2021repurposing,zhang2021datasetgan}. Various approaches have been proposed for few-shot part segmentation, such as Rep-GAN~\cite{tritrong2021repurposing}, DatasetGAN~\cite{zhang2021datasetgan}, and DDPM~\cite{baranchuk2021label}. However, existing few-shot part segmentation methods usually extract pretrained features from the generative models and train a pixel-wise classifier over the pretrained features, thereby failing to learn task-specific features for part segmentation. In this work, we design a fine-tuning strategy to progressively tuning an image generator into a segmentation generator, which leverages the GAN priors by reducing the gap between the pre-train task (\textit{i.e.}, image generation) and the downstream task (\textit{i.e.}, part segmentation) in a ``pre-training''-``fine-tuning'' paradigm.

\section{Approach}
In this section, we introduce our proposed Progressive Fine-Tuned GAN (PFTGAN) for few-shot part segmentation. Our proposed PFTGAN can learn task-specific features in a ``pre-training''-``fine-tuning'' paradigm and reduce both ``task gap'' and ``domain gap'' to leverage GAN priors. 

Figure~\ref{fig:framework} is an overview of our framework, which illustrates our two-stream network architecture, and optimization strategies of the two streams.
Figure~\ref{fig:framework} (A) shows the architecture of PFTGAN, which consists of an encoder and a two-stream decoder. Given an input image, we first feed it into a pre-trained encoder to obtain its corresponding latent code. The encoder is trained by using the GAN inversion techniques, hence a StyleGAN generator can take the latent code as input and recover the input image. The decoder consists of a segmentation stream and an image stream. These two streams are supervised by RGB segmentation maps and original images, respectively, where the layers corresponding to coarse spatial resolutions (\textit{e.g.}, $4\times 4$ to $32\times 32$) are shared. Both streams are initialized by pre-trained StyleGAN weights. Figure~\ref{fig:framework} (B) is an illustration of our proposed progressive fine-tuning for the segmentation stream. We first fix parameters of the convolutional layers and fine-tune the ``toRGB'' layers by interpolated images. After that, we set convolutional layers trainable and fine-tune the decoder by interpolated images. Finally, we further fine-tune the ``toRGB'' layers by RGB segmentation maps. Projecting part segmentation maps into the RGB space reduces the ``task gap'' of generation and segmentation; and interpolating RGB segmentation maps and original images further reduces the ``domain gap''. Figure~\ref{fig:framework} (C) shows the optimization of the image stream, which can be regarded as a self-supervised auto-encoder and is trained on large-scale support (unlabeled) images. After completing the fine-tuning process, we train a MLP classifier on top of the StyleGAN’s pixel-wise feature vectors to match part segmentation labels following~\cite{tritrong2021repurposing,zhang2021datasetgan}.

\subsection{Network Architecture}
As shown in Figure~\ref{fig:framework} (A), we propose a two-stream architecture (\textit{i.e.}, a segmentation stream and an image stream), where the segmentation stream can generate task-specific features for part segmentation and the image stream can provide spatial constraints. StyleGAN~\cite{karras2019style} shows that in a generator, the layers corresponding to coarse spatial resolutions (\textit{e.g.}, $4 \times 4$ to $32 \times 32$) dominate high-level and spatial information of a generated image, and the layers corresponding to finer spatial resolutions (\textit{e.g.}, $32 \times 32$ to $1024 \times 1024$) dominate low-level texture or color information. Inspired by this property, we keep the layers corresponding to coarse spatial resolutions shared in the two streams. Such a design can provide spatial constraints and benefit the fine-tuning process from two aspects: 1) the downstream task of the image stream has less gap to the pre-train task of image generation than the segmentation stream, and thus GAN priors can be better preserved. 2) The optimization of image stream can be regarded as learning a self-supervised auto-encoder, enabling our model to benefit from large-scale support images (\textit{e.g.}, unlabeled images from the same dataset).

We formulate the optimization of the two-stream decoder as a multi-task learning problem. Specifically, we denote an input image as $\mathbf{X} \in \mathbb{R}^{H \times W \times 3}$ and its corresponding part segmentation label as $\mathbf{Y} \in \mathbb{R}^{H \times W \times \{1, ..., K\}}$. To reduce the ``task gap'' of image generation and part segmentation, we project the part segmentation label $\mathbf{Y}$ into the RGB space, which is denoted as $\mathbf{M} \in \mathbb{R}^{H \times W \times 3}$. We first use a pre-trained encoder to obtain the corresponding latent code:
\begin{equation}\label{eqn:encode}
\mathbf{w} = \mathcal{E}(\mathbf{X}),
\end{equation}
where $\mathcal{E}$ is the encoder, $\mathbf{w} \in \mathbb{R}^{N \times C}$ is the obtained latent code, $N$ and $C$ are the layer number and latent space dimension of the StyleGAN generator, respectively. The two-stream decoder takes as input such a latent code and aims to generate RGB segmentation maps and recover the input image, respectively:
\begin{equation}\label{eqn:decode}
\mathbf{\hat{M}} = \mathcal{G}_{seg}(\mathbf{w}), \quad \mathbf{\hat{X}} = \mathcal{G}_{img}(\mathbf{w}),
\end{equation}
where $\mathcal{G}_{seg}$ and $\mathcal{G}_{img}$ are the segmentation stream decoder (generator) and the image stream decoder, respectively. We follow JoJoGAN~\cite{chong2021jojogan} to use perceptual loss (\textit{i.e.}, lpips) to optimize the two-stream decoder:
\begin{equation}\label{eqn:lpips}
    \hat {\theta } = \underset{\theta}{\arg\min}( \mathcal {L}_{lpips}(\mathbf{\hat{M}}, \mathbf{M}) + \mathcal {L}_{lpips}(\mathbf{\hat{X}}, \mathbf{X})),
\end{equation}
where $\mathcal {L}_{lpips}$ is the lpips loss, and $\theta$ denotes the parameters of the decoder. Note that optimization of the image stream can be regarded as learning a self-supervised auto-encoder, which dose not need part annotations. Thus we can leverage the large-scale support images as shown in Figure~\ref{fig:framework} (C). Note that theoretically, the support image can be any unlabeled images, \textit{e.g.}, the images used in StyleGAN pre-training. In our experiments, we find that it is effective to adopt the testing images (without any label). The training objective can be formulated as:
\begin{equation}\label{eqn:lpips_aux}
    \hat {\theta } = \underset{\theta}{\arg\min} \mathcal {L}_{lpips}(\mathbf{\hat{M}}, \mathbf{M}) + ( \mathcal {L}_{lpips}(\mathbf{\hat{X}}, \mathbf{X}) +  \frac{1}{N} {\textstyle \sum_{i}} \mathcal {L}_{lpips}(\mathbf{\hat{X}}'_i, \mathbf{X}'_i)) ,
\end{equation}
where $N$ is the number of support images, $\mathbf{X}'_i$ and $\mathbf{\hat{X}}'_i$ represent the $i^{th}$ support images and the corresponding outputs of the image stream, respectively.

\begin{algorithm}[t]
\caption{The progressive fine-tuning strategy}
\label{alg:algorithm}

\begin{algorithmic}[1] 
\STATE \textbf{Input}: Training image $\mathbf{X}$; part label $\mathbf{Y}$; support images $\mathbf{X}'$; maximal iteration number $T$; pre-trained encoder $\mathcal{E}$; StyleGAN-initialized two-stream decoder $\{\mathcal{G}_{img},\mathcal{G}_{seg}\}$.
\STATE Encode the image into the latent code space: $\mathcal{E}: \mathbf{X} \to \mathbf{w}$.
\STATE Project the segmentation label $\mathbf{Y}$ into the RGB space to obtain the RGB segmentation maps $\mathbf{M}$.
\STATE Interpolate the RGB segmentation maps $\mathbf{M}$ and the original image $\mathbf{X}$ to obtain the interpolated image $\mathbf{X}_\mathbf{M}$.
\FOR{i=1, 2, $\cdots$, T}
\STATE Finetune the ``toRGB'' layers in $\mathcal{G}_{seg}$ using $\mathbf{X}_\mathbf{M}$;
\STATE Finetune the ``toRGB'' layers in $\mathcal{G}_{img}$ using $\mathbf{X}'$.
\ENDFOR
\FOR{i=1, 2, $\cdots$, T}
\STATE Finetune all the parameters in $\mathcal{G}_{seg}$ using $\mathbf{X}_\mathbf{M}$;
\STATE Finetune all the parameters in $\mathcal{G}_{img}$ using $\mathbf{X}'$.
\ENDFOR
\FOR{i=1, 2, $\cdots$, T}
\STATE Finetune the ``toRGB'' layers in $\mathcal{G}_{seg}$ using $\mathbf{M}$;
\STATE Finetune the ``toRGB'' layers in $\mathcal{G}_{img}$ using $\mathbf{X}'$.
\ENDFOR
\STATE \textbf{Output}: Optimized parameters of  $\{\mathcal{G}_{img}, \mathcal{G}_{seg}\}$.
\end{algorithmic}
\end{algorithm}

\subsection{Progressive Fine-tuning Strategy}
It is non-trivial to fine-tune the decoder by using only a few samples, and directly optimizing the decoder to fit the part labels would lead to severe over-fitting. As mentioned above, we project part labels into RGB segmentation maps, which can significantly reduce the task gap between image generation and part segmentation. However, there are still domain gaps between RGB segmentation maps and original images, which would also deteriorate GAN priors during the fine-tuning process. Moreover, the trainable parameters in an StyleGAN generator can be roughly grouped as convolutional parameters and ``toRGB'' ones (\textit{i.e.}, $\theta = \{\theta_{convs}, \theta_{toRGB}\}$), and it is straightforward to fine-tune only part of parameters to alleviate over-fitting. To this end, we propose a progressive fine-tuning strategy, which is shown in Figure~\ref{fig:framework} (B).

There are three steps in our progressive fine-tuning strategy.

\textbf{Step 1}: Fine-tune the ``toRGB'' layers using interpolated images. We interpolate the RGB segmentation map $\mathbf{M}$ and the original image $\mathbf{X}$ into an interpolated image (shown in Figure~\ref{fig:framework} (B)), \textit{i.e.},
\begin{equation}\label{eqn:interp}
\mathbf{X}_\mathbf{M} = \lambda\mathbf{X} + (1-\lambda)\mathbf{M},
\end{equation}
where $\mathbf{X}_\mathbf{M}$ is the interpolated image, $\lambda$ is a hyper-parameter to control the ratio of the two components. We optimize the ``toRGB'' layers using the following objective functions:
\begin{equation}\label{eqn:step1}
    \hat {\theta}_{toRGB} = \underset{\theta_{toRGB}}{\arg\min} \ \mathcal{L}_{lpips}(\mathbf{\hat{M}}, \mathbf{X}_\mathbf{M}),
\end{equation}
where ${\theta}_{toRGB}$ denotes the parameters of ``toRGB'' layers.

\textbf{Step 2}: Fine-tune both the ``toRGB'' layers and convolutional layers using interpolated images. In this step, we set the convolutional layers to be trainable to further obtain task-specific features. Since we have fine-tuned the ``toRGB'' layers in Step 1, the gradients over convolutional layers would be moderate. The segmentation stream is optimized according to:
\begin{equation}\label{eqn:step2}
    \{\hat {\theta}_{toRGB}, \hat {\theta}_{convs}\} = \hat {\theta} = \underset{\theta}{\arg\min} \ \mathcal{L}_{lpips}(\mathbf{\hat{M}}, \mathbf{X}_\mathbf{M}).
\end{equation}

\textbf{Step 3}: Fine-tune the ``toRGB'' layers using RGB segmentation maps. Finally, we optimize the ``toRGB'' layers to fit RGB segmentation maps, and the outputs can facilitate the training of the subsequent Pixel-level Classifier (which will be depicted later). Such an optimization can be denoted as:
\begin{equation}\label{eqn:step3}
    \hat {\theta}_{toRGB} = \underset{\theta_{toRGB}}{\arg\min} \ \mathcal{L}_{lpips}(\mathbf{\hat{M}}, \mathbf{M}).
\end{equation}

\textbf{Clarification of different optimization equations:} Eq.~(\ref{eqn:lpips}) and  Eq.~(\ref{eqn:lpips_aux}) are the overall formulations to optimize our PFTGAN, and Eq.~(\ref{eqn:step1}) to ~(\ref{eqn:step3}) show the detailed loss functions over the segmentation stream in three fine-tuning steps. More details can be found in Algorithm~\ref{alg:algorithm}, which summarizes the main process of our proposed progressive fine-tuning strategy.

\subsection{Pixel-level Classifier}
After the progressive fine-tuning, the optimized model is used to extract pixel-level representations of the input image. In particular, we follow~\cite{tritrong2021repurposing,zhang2021datasetgan} to extract the activation map from every layer (or some subset of layers) of the segmentation stream, and obtain $\{S_0, S_1.., S_k\}$, where the dimensions of $S_i$ are $\{H_i, W_i, C_i\}$. The extracted representations from all layers are upsampled to the image size and concatenated, and this results in feature vectors for all pixels of the labeled images. 
Specifically, the size of $S_i$ can be spatially interpolated to input size $H \times W \times C_i$. The resulting pixel-level representation is a $C$ dimensional feature vector, where $C= {\textstyle \sum_{i}^{k}} C_i$. 
This process maps each three dimensional RGB pixel to a $C$ dimensional feature vector. 
After that, we train a multi-layer perception (MLP) over these pixel-level representations, which aim to predict a semantic label of each pixel available for training images. We adopt the training settings from DatasetGAN~\cite{zhang2021datasetgan} and utilize them for all other methods in our experiments.


\begin{figure*}[t]
  \centering
  \includegraphics[width=1\linewidth]{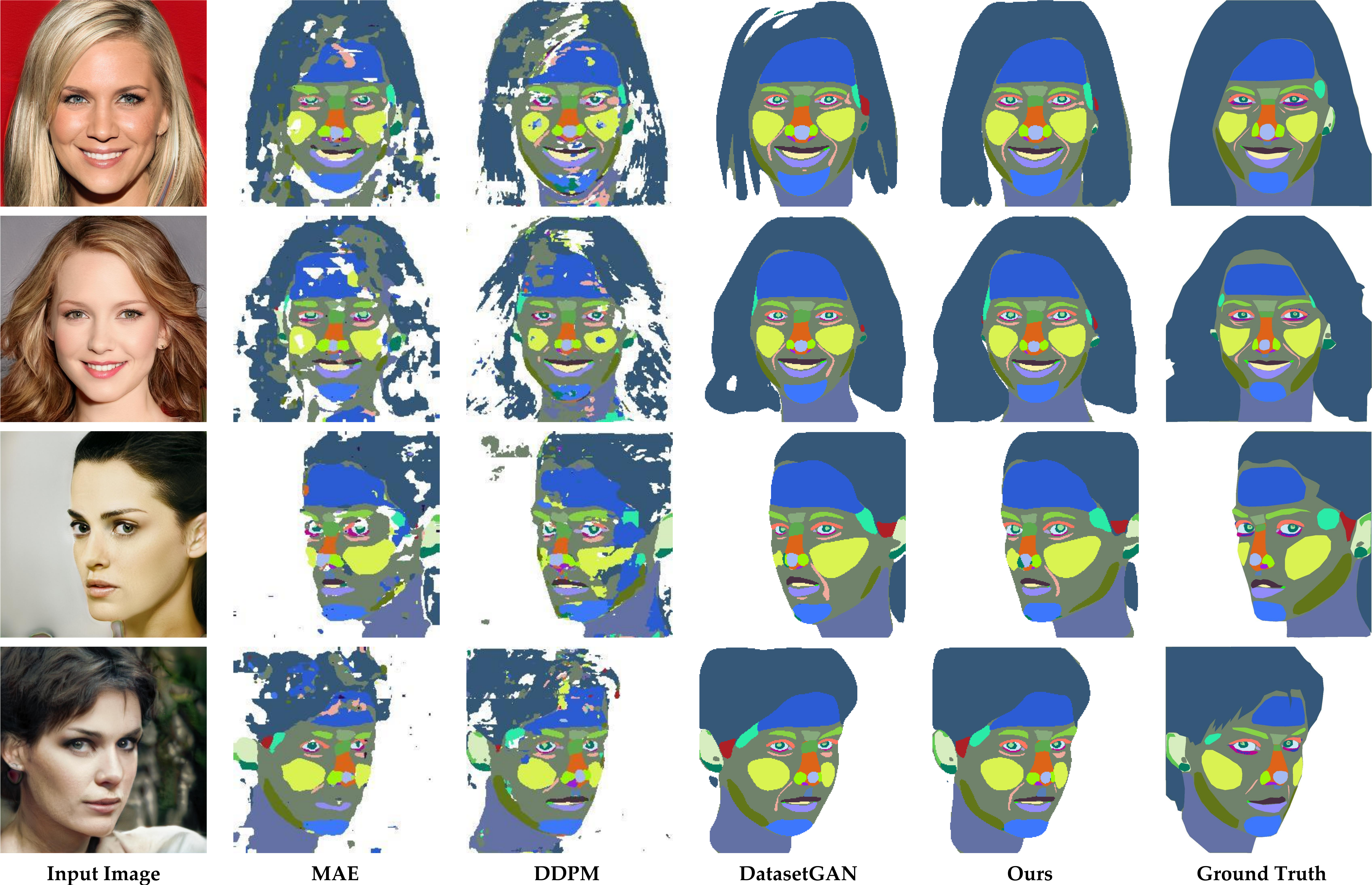}
  \caption{Qualitative comparison on the FFHQ-34 dataset. Our model can better avoid noises (compared to MAE~\cite{he2021masked} and DDPM~\cite{baranchuk2021label}) and achieve more precise spatial alignment (compared DatasetGAN~\cite{zhang2021datasetgan}).}
  \label{fig:ffhq}
\end{figure*}

\section{Experiment}

\subsection{Experiment setup}

In this section, we present the experimental details of our approach. We outline the datasets, implementation details, baseline methods and evaluation metrics. Code is available at \url{https://github.com/hanmengya1996/PFTGAN}.

\textbf{Datasets}. FFHQ-34 and LSUN CAR-20 are widely used for few-shot part segmentation, which are collected by DatasetGAN~\cite{zhang2021datasetgan} and DDPM~\cite{baranchuk2021label}. Note that the numbers in the dataset name correspond to the number of semantic/part classes. 


  

\textbf{Implementation details}. We use StyleGAN~\cite{karras2019style} as the generator, and keep the layers corresponding to coarse spatial resolutions shared in two streams. Both streams are initialized by pre-trained StyleGAN weights.
Our training pipeline starts by progressively fine-tuning a pre-trained GAN on a few labeled images. Then, the optimized model is used to extract pixel-level representations of the input images. 
Finally, we train a multi-layer perception (MLP), which aim to take these pixel-level representations as input and predict a semantic label of each pixel available for images. The MLP was trained with a cross-entropy loss and the initial learning rate of 0.5 using SGD optimizer. In the experiment, we set $\lambda$ to be 0.1.

\textbf{Baselines}. We compare to three baseline methods that are based on different pre-trained features, \textit{i.e.}, GAN-based features, Diffusion-based features, and Self-supervised Learning (SSL-based) features. We compare to them because they are the most competitive ones:

\begin{itemize}[leftmargin=*]
\item DatasetGAN~\cite{zhang2021datasetgan} — this method exploits the pixel-level features produced by pre-trained GANs. The extracted representations from all layers of the pre-trained styleGAN are upsampled to the image size and concatenated, forming the feature vectors for few-shot part segmentation.

\item DDPM~\cite{baranchuk2021label} - this method exploits the pixel-level representations produced by diffusion models. The extracted representations from all blocks of the UNet decoder are upsampled to the image size and concatenated, forming the feature vectors for few-shot part segmentation.

\item MAE~\cite{he2021masked} — one of the state-of-the-art self-supervised methods, which learns an autoencoder to reconstruct missing patches. We follow previous work~\cite{baranchuk2021label} to first train a ViT-Large~\cite{dosovitskiy2020image} by MAE, and then extract ViT features for part segmentation. 
\end{itemize}

\textbf{Evaluation Metric}. To evaluate the segmentation performance of the segmentation network, Mean intersection over union (mIoU) was employed by comparing the predicted segmentation masks to the ground truth ones which were annotated manually.
\begin{equation}
    mIOU = {\textstyle \sum_{i}} \frac{TP_i / (TP_i + FP_i +FN_i)}{n},
\end{equation}
where $TP_i$ is the number of correctly classified pixels of class $i$, $FP_i$ is the number of wrongly classified pixels of class $i$, $FN_i$ is the number of wrongly misclassified pixels of class $i$, and $n$ is the number of semantic classes.



\subsection{Comparisons}
We evaluate few-shot part segmentation performance on two widely obtained datasets: FFHQ-34 and LSUN CAR-20.

\textbf{Experiments on FFHQ-34}.
We first compare our approach to the state-of-the-art methods on the FFHQ-34 (real image) dataset in mIoU metric. The results are summarized in Table~\ref{tab:ffhq}. Our model outperforms the state-of-the-art methods in all settings. It is worth noting that our approach obtains more improvement in 1-shot task. In the 1-shot part segmentation task, we obtain the highest score of 0.380 mIOU. Compared with three baselines, DatasetGAN, DDPM,and, MAE, our model achieves significant improvement of 2\%, 9.5\%, and 13.4\%, respectively. In the 5-shot part segmentation task, our approach outperforms state-of-the-art methods in this dataset with mIoU increases of 3.9\%, 0.3\%, and 2.7\% respectively for DatasetGAN, DDPM, and MAE. This improvement mainly derives from the designed two-stream decoder and progressive fine-tuning strategy, which can effectively reduce the task gap and domain gap of image generation and part segmentation.

We also compare our approach to the state-of-the-art methods on the GAN generated images. The results are summarized in Table~\ref{tab:gen}.  Compared with the state-of-the-art methods, our method obtains the best performance on the GAN generated images in all settings, except the 5-shot task. In the 1-shot part segmentation task, our model achieves significant improvement of 4.5\%, 8.6\%, and 10.5\% compared to DatasetGAN, DDPM, and MAE, respectively. Qualitative comparisons (2-shot) are shown in Figure~\ref{fig:ffhq}.

\begin{figure}[t]
  \centering
  \includegraphics[width=\linewidth]{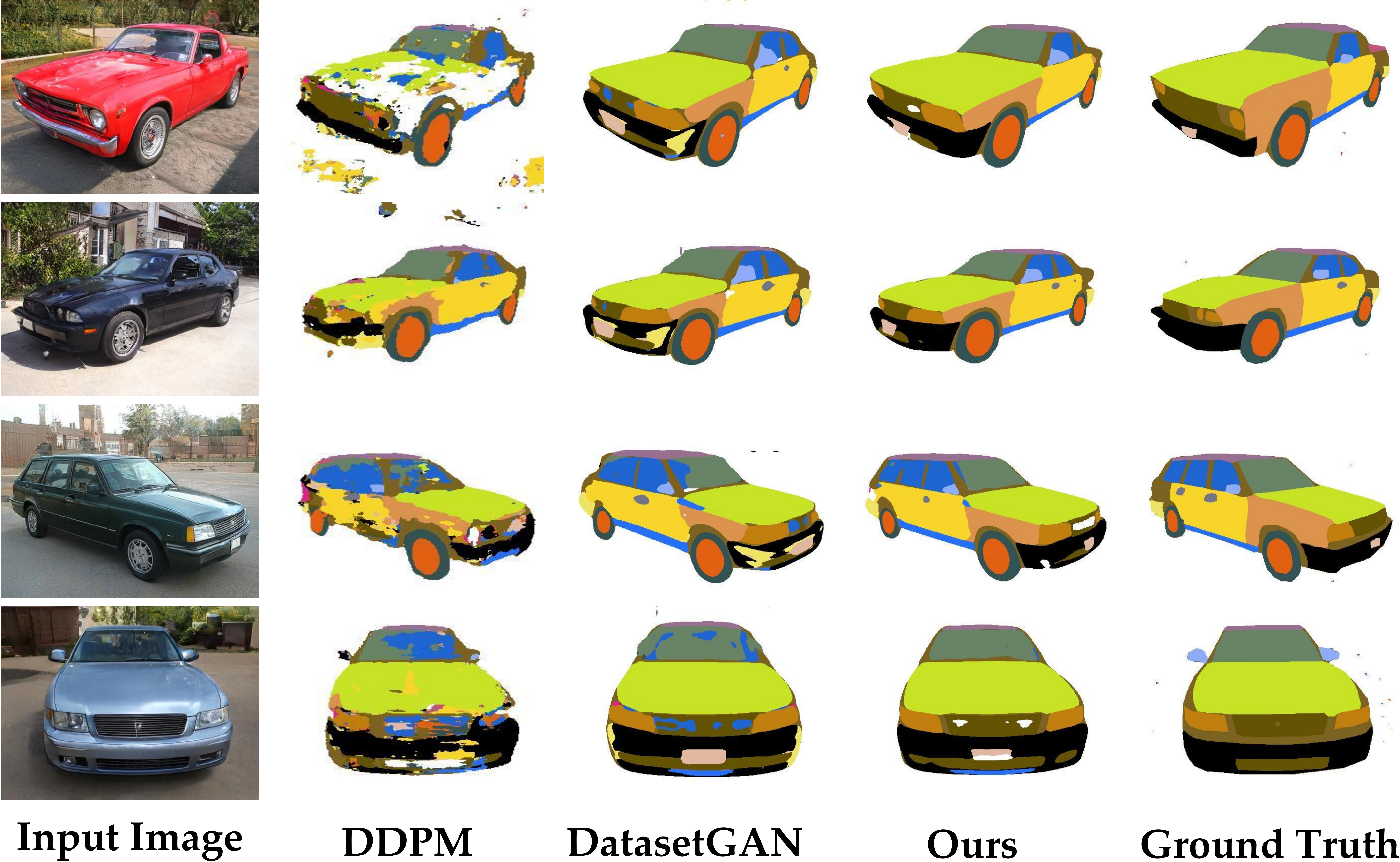}
  \caption{Qualitative comparison on the LSUN CAR-20 dataset. Our model can better avoid noises and generate consistent semantic parts.}
  \label{fig:car}
\end{figure}

\begin{table}
{\small
\caption{Comparison results on the FFHQ-34 dataset (real images) in terms of mIOU.}
\label{tab:ffhq}
\begin{center}
\resizebox{0.95\columnwidth}{!}{
    \begin{tabular}{c|c|c|c|c|c}
    \hline
    Methods & 1-shot & 2-shot & 3-shot & 4-shot & 5-shot\\
    \hline \hline
    DatasetGAN~\cite{zhang2021datasetgan}  & 0.360 & 0.430 & 0.436 & 0.439 & 0.439 \\
    \hline
    DDPM~\cite{baranchuk2021label} &  0.285 & 0.412 & 0.436 & 0.460 & 0.473\\
    \hline
    MAE~\cite{he2021masked} &  0.246 & 0.379 & 0.425 & 0.437 & 0.451\\
    \hline
    Ours & \textbf{0.380} & \textbf{0.457} & \textbf{0.467} & \textbf{0.472} &\textbf{0.478} \\
    \hline
    \end{tabular}
    }
\end{center}
}
\end{table}

\begin{table}
{\small
\caption{Comparison results on the FFHQ-34 dataset (GAN generated images) in terms of mIOU.}
\label{tab:gen}
\begin{center}
\resizebox{0.95\columnwidth}{!}{
    \begin{tabular}{c|c|c|c|c|c}
    \hline
    Methods & 1-shot & 2-shot & 3-shot & 4-shot & 5-shot\\
    \hline \hline
    DatasetGAN~\cite{zhang2021datasetgan}  & 0.384 & 0.404 & 0.462 & 0.466 &  0.464\\
    \hline
    DDPM~\cite{baranchuk2021label} & 0.343  & 0.399 & 0.468  & 0.489 & \textbf{0.513}\\
    \hline
    MAE~\cite{he2021masked} & 0.324  & 0.390 & 0.439 & 0.466 & 0.490\\
    \hline
    Ours & \textbf{0.429} & \textbf{0.438} & \textbf{0.491} & \textbf{0.499} & 0.507 \\
    \hline
    \end{tabular}
    }
\end{center}
}
\end{table}

\begin{table}
{\small
\caption{Comparison results on the LSUN CAR-20 dataset in terms of mIOU.}
\label{tab:car}
\begin{center}
\resizebox{0.95\columnwidth}{!}{
    \begin{tabular}{c|c|c|c|c|c}
    \hline
    Methods & 1-shot & 2-shot & 3-shot & 4-shot & 5-shot\\
    \hline \hline
    DatasetGAN~\cite{zhang2021datasetgan}  & 0.270 & 0.302 & 0.295 & 0.303 & 0.304\\
    \hline
    DDPM~\cite{baranchuk2021label} &  0.251 & 0.306 & 0.311 & 0.321 & 0.298 \\
    \hline
    Ours & \textbf{0.338}& \textbf{0.373} & \textbf{0.372}& \textbf{0.393} & \textbf{0.389}  \\
    \hline
    \end{tabular}
    }
\end{center}
}
\end{table}

\begin{figure}[!t]
  \centering
  \includegraphics[width=\linewidth]{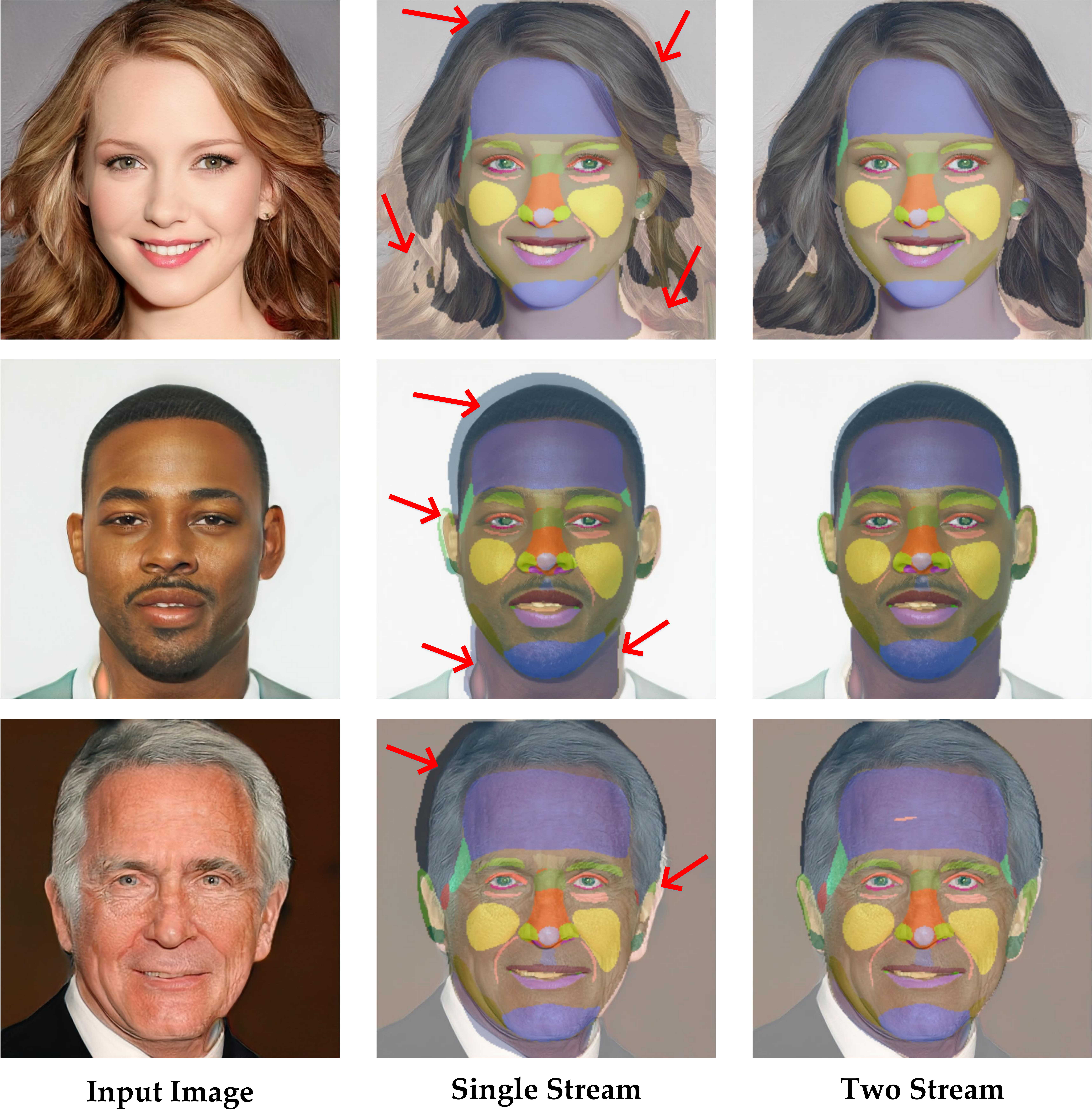}
  \caption{Qualitative analysis of the proposed two-stream architecture. The image stream can provide spatial constraints. Red arrows show the spatial misalignment.}
  \label{fig:two_stream}
\end{figure}

\textbf{Experiments on LSUN CAR-20}.
We also evaluate our model on the LSUN CAR-20 dataset, and the results in Table~\ref{tab:car} show the effectiveness of our model. In the 1-shot setting, our method shows a significant improvement over existing methods~\cite{zhang2021datasetgan,baranchuk2021label}; it delivers a gain of 6.8\% and 7.2\% in mIoU compared with DatasetGAN~\cite{zhang2021datasetgan} and DDPM~\cite{baranchuk2021label},
respectively. Similarly, in the other settings, our method also outperforms DatasetGAN and DDPM by a large margin. This is due to the fact that PFTGAN is capable of learning task-specific features for part segmentation by leveraging GAN priors.
Figure~\ref{fig:car} shows the qualitative results (2-shot) from the three methods. We see that, compared to DatasetGAN~\cite{zhang2021datasetgan} and DDPM~\cite{baranchuk2021label}, the proposed approach produces more accurate segmentation masks.

\begin{figure*}[t]
  \centering
  \includegraphics[width=0.9\linewidth]{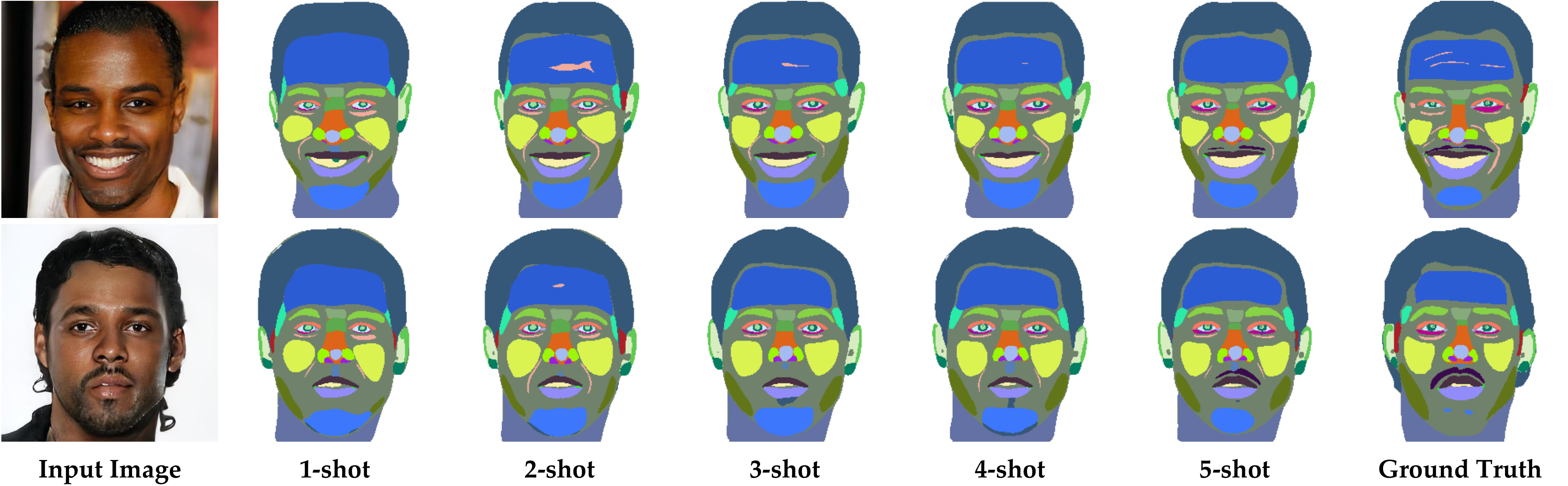}
  \caption{Qualitative results on FFHQ-34. Our model can generate reasonable results given only one training sample, and rich parts can be obtained once appeared in the training samples (\textit{e.g.}, only the fifth sample contains ``beard'').}
  \label{fig:shotnumber}
\end{figure*}

\begin{table}
{\small
\caption{Ablation study of the proposed two-stream architecture in terms of mIOU.}
\label{tab:multi-task}
\begin{center}
\resizebox{0.65\columnwidth}{!}{
    \begin{tabular}{c|c|c}
    \hline
    Method & FFHQ-34 & LSUN CAR-20\\
    \hline \hline
     Baseline~\cite{zhang2021datasetgan} &  0.384 & 0.270\\
     \hline
    Single Stream & 0.379 & 0.274\\
    \hline
    Two Stream & \textbf{0.429} &\textbf{0.338}\\
    \hline
    \end{tabular}
    }
\end{center}
}
\end{table}

\begin{table}
{\small
\caption{Ablation study of the desidned ``shared stream'' in terms of mIOU.}
\label{tab:multi-task}
\begin{center}
\resizebox{0.55\columnwidth}{!}{
    \begin{tabular}{c|c}
    \hline
    Method & FFHQ-34 \\
    \hline \hline
    w/o ``shared stream''  &  0.383\\
     \hline
    ``shared stream'' & \textbf{0.429} \\
    \hline
    \end{tabular}
    }
\end{center}
}
\end{table}

\subsection{Ablation Studies}

\textbf{Two-stream architecture}. The proposed framework is a two-stream architecture, including a segmentation stream and an image stream.
To verify the effectiveness of two-stream architecture, we conduct ablation studies on the GAN generated images. The ablation analysis results are tabulated in Table~\ref{tab:multi-task}. From the results in Table~\ref{tab:multi-task}, it is evident that two-stream brings an improvement over the baseline. Note that directly fine-tuning the pre-trained GAN to generate an RGB segmentation map under the single-stream architecture degrades the segmentation performance due to ``task gap'' and ``domain gap''. Under the two-stream architecture, our method achieves the best performance on all datasets in the 1-shot task. This improvements derive from the segmentation stream can generate task-specific features for part segmentation and the image stream can provide spatial constraints simultaneously. We visualize and compare the generated RGB segmentation map from the single-stream architecture and our proposed two-stream architecture in Figure~\ref{fig:two_stream}, which could intuitively explain why the two-stream outperforms the single-stream by such a large margin.

We also conducted an ablation study to evaluate the effectiveness of the ``shared stream". The experimental results are as follows. The ``shared stream" design can provide spatial constraints in the segmentation stream and improve the performance.

\begin{table}
{\small
\caption{Ablation study of our proposed progressive fine-tuning strategy in terms of mIOU. ``Vanilla ft.'' indicates directly fine-tuning StyleGAN by RGB segmentation maps.}
\label{tab:ft}
\begin{center}
\resizebox{0.8\columnwidth}{!}{
    \begin{tabular}{l|c|c}
    \hline
    Method & FFHQ-34 & LSUN CAR-20\\
    \hline \hline
     Baseline~\cite{zhang2021datasetgan} & 0.384 & 0.270\\
     \hline
    Vanilla ft.& 0.347 & 0.267\\
     \hline
    Step 1& 0.385 & 0.281\\
    Step 1 + Step2&  0.424 & 0.327\\
   Step 1 + Step2 + Step3& \textbf{0.429} &\textbf{0.338}\\
    \hline
    \end{tabular}
    }
\end{center}
}
\end{table}



\textbf{Progressive Fine-tuning strategy}.
To verify the effectiveness of the proposed progressive fine-tuning strategy, we conduct ablation experiments regarding the fine-tuning strategy. The results are given in Table~\ref{tab:ft}. Without the fine-tuning, the baseline obtain the worst performance. Step 1, Step2, together with Step 3, brings the best performance, reflecting that the three fine-tuning steps could cooperate together for further performance improvement.

\textbf{Shot number}.
We perform experiments under the 5 settings that vary the number of examples with part annotations (1-shot, 2-shot, 3-shot, 4-shot, 5-shot). Table~\ref{tab:gen} shows mIOU of the various shot number on face segmentation.  Quantitative results are shown in Figure~\ref{fig:shotnumber}. From the visualization, it is obvious that the segmentation performance gets improved gradually when the labeled image number increases from 1 to 5.

\begin{figure}
  \centering
  \includegraphics[width=0.95\linewidth]{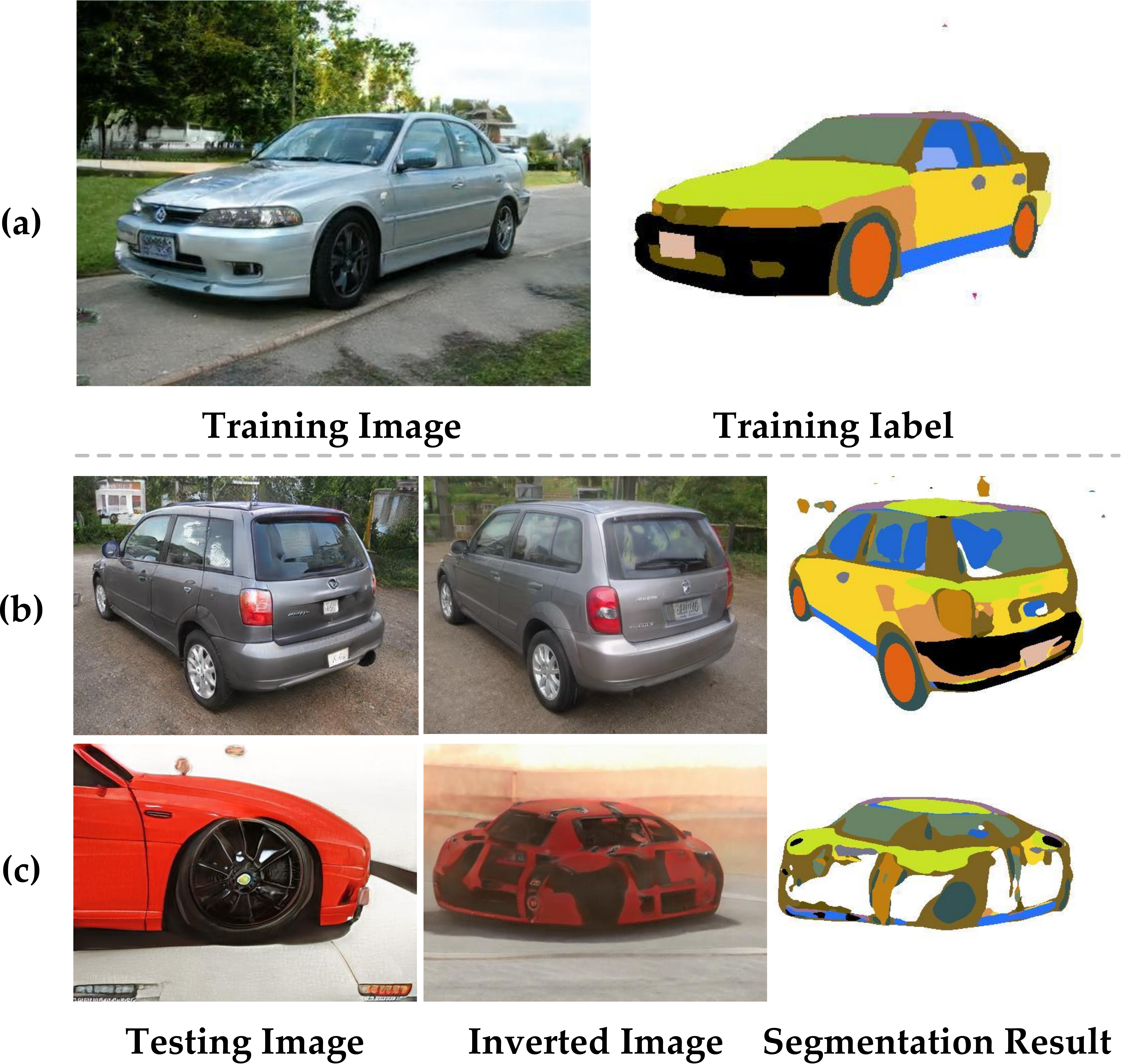}
  \caption{Failure cases are mainly two-fold, \textit{i.e.}, (b) large view variance with unseen part, and (c) out-of-domain cases that the GAN inversion model cannot invert it properly.}
  \label{fig:fail}
\end{figure}

\section{Conclusion}

In this work, we propose a Progressive Fine-Tuned GAN (PFTGAN) to learn task-specific features for few-shot part segmentation. Our proposed method can effectively reduce the task gap and domain gap between image generation and part segmentation. Moreover, we design a two-stream architecture, where the self-supervised auto-encoder enables our model to benefit from large-scale support images and facilitates to preserve the spatial information. Extensive experiments demonstrate the effective of our proposed framework. Nevertheless, our method is limited in two main aspects: 1) our model relies on the result of GAN inversion models (\textit{e.g.}, Figure~\ref{fig:fail} (c)); and 2) diffusion models can outperform GAN-based models when more training data (\textit{e.g.}, more than 5-shot) are provided. In the future, we intend to: 1) design a better GAN inversion model for part segmentation; 2) improve the performance by integrating with diffusion-based and SSL-based (self-supervised learning) part segmentation; 3) extend our work for more complex datasets that contain multiple classes (\textit{e.g.}, ImageNet).

\section*{ACKNOWLEDGMENTS}
This work was supported by Science and Technology Innovation 2030 –“Brain Science and Brain-like Research” Major Project (No. 2021ZD0201405), the Special Fund of Hubei Luojia Laboratory under Grant 220100014, the National Natural Science Foundation of China under Grant 62141112, and the Science and Technology Major Project of Hubei Province (Next-Generation AI Technologies) under Grant 2019AEA170.
  

\end{document}